\documentclass[letterpaper, 10 pt, conference]{ieeeconf}  

\IEEEoverridecommandlockouts                              

\usepackage{caption}
\usepackage{graphicx} 
\usepackage{amsmath} 
\usepackage{amssymb}  
\usepackage{subcaption}

\title{\LARGE \bf
BoxTwin: Learning Elastoplastic Articulated Object Dynamics\\ from Videos
}

\author{Heng Zhang*$^{1}$, Gehan Zheng*$^{1}$, Kaifeng Zhang$^{1}$, Jay Song$^{1}$, Shivansh Patel$^{1}$, \\ Sonny Hu$^{1}$, Yunzhu Li$^{1}$, Changxi Zheng$^{1}$ and Peter Yichen Chen$^{2}$
\thanks{*\ Equal contribution}
\thanks{$^{1}$Heng Zhang, Gehan Zheng, Kaifeng Zhang, Jay Song, Shivansh Patel, Sonny Hu, Yunzhu Li and Changxi Zheng are with SceniX, Inc.}%
\thanks{$^{2}$Peter Yichen Chen is with the University of British Columbia.}%
}

\begin{document}


\maketitle

\thispagestyle{empty}
\pagestyle{empty}

\begin{abstract}
Digital twins enable robots to anticipate and adapt to physical interactions, but existing models struggle with elastoplastic articulated objects (EAOs) that exhibit nonlinear elasticity, plastic yielding, and damage accumulation. We present BoxTwin, an interactive digital‑twin framework that learns the full dynamics of EAOs from videos. Our pipeline reconstructs the scene, identifies a physics‑aware constitutive model for each EAO. Experiments on manual folding and dual‑arm manipulation of EAOs show that BoxTwin accurately tracks joint trajectories and reproduces post‑contact plastic behavior over long horizons. By integrating video‑driven reconstruction with elastoplastic‑damage modeling, BoxTwin advances digital twins toward predictive, adaptive control of deformable articulated objects in unstructured environments.
\end{abstract}

\section{INTRODUCTION}

Integrating digital twins into robotic platforms heralds a new era of safer, more adaptable human‑robot collaboration within complex physical settings\cite{jiang2025phystwin, malik2021digital}. By furnishing physics-based real-time replicas of tangible objects, digital twins allow robots to foresee interactions, refine manipulation tactics, and improve decision making amid unstructured environments\cite{mu2025robotwin}, abilities that are especially vital when handling deformable items. However, modeling deformable objects persists as a major obstacle in robotic manipulation\cite{feng2024pie, jiang2024vr, xie2024physgaussian, chen2022virtual, li2023pac, qiao2022neuphysics, zhang2024physdreamer, zhong2024reconstruction}; current techniques often do not cover the full range of mechanical responses elicited during contact. This issue becomes pronounced for articulated elastoplastic objects, which combine reversible elastic deformations with irreversible plastic changes and feature joints that enable intricate motions (e.g. boxes, doors, hinges, sheet‑metal assemblies). Unlike rigid or purely elastic bodies, these objects display highly non-linear dynamics that evolve throughout interaction, posing considerable challenges for precise simulation and control.
\begin{figure}[htbp]
    \centering
    \includegraphics[width=\linewidth]{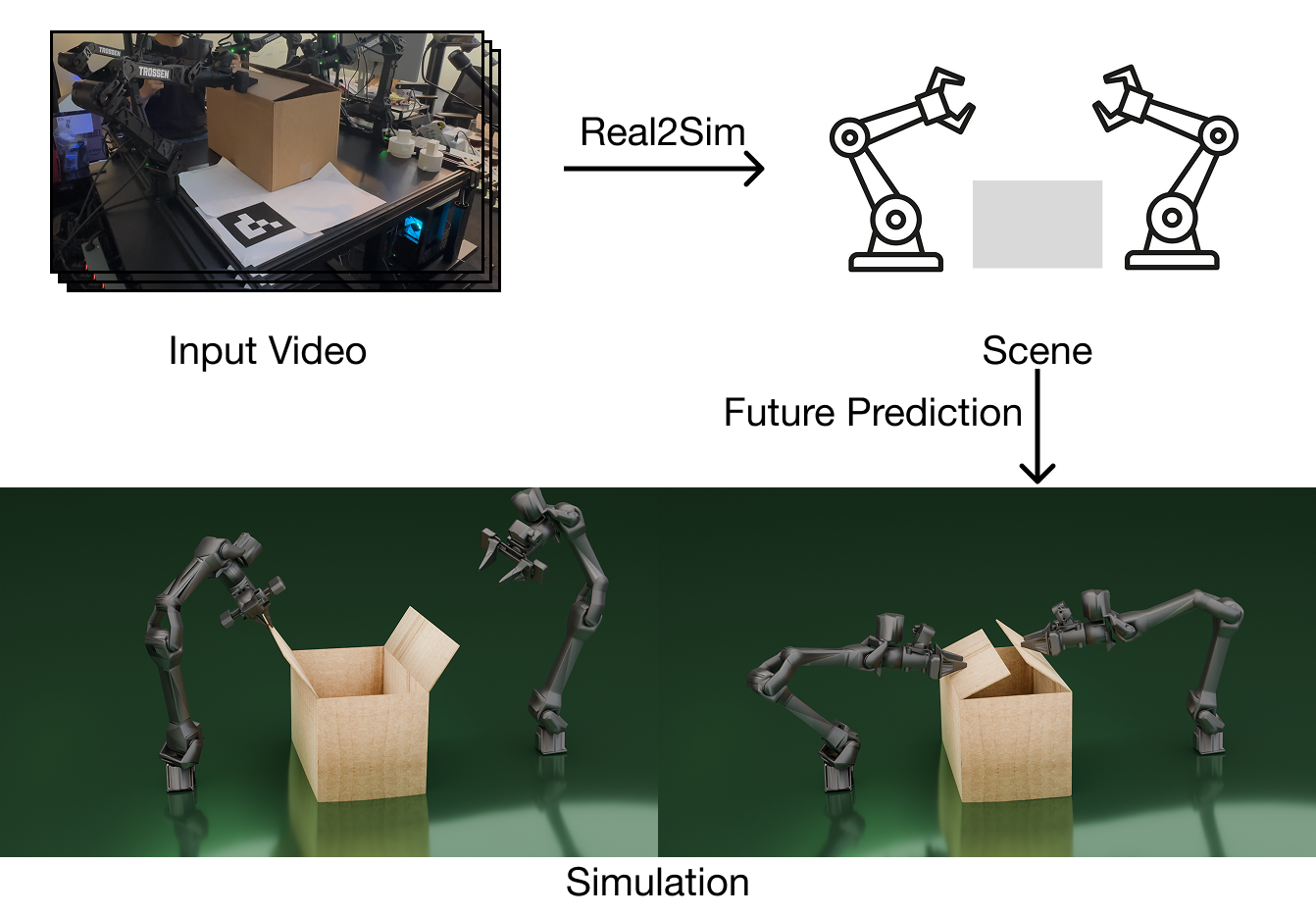} 
    \caption{Our workflow first reconstructs the full scene from the input video, extracting the EAOs’ parameters; with this reconstructed model, we can subsequently predict future states.}
    \label{fig:teaser}
    \vspace{-9mm}
\end{figure}

The difficulty arises from three tightly coupled physical effects. First, non-linear elasticity produces stress-strain relationships far from Hookean behavior, leading to energy dissipation and interdegree-of-freedom coupling that a linear model cannot capture. Second, plastic deformation introduces permanent shape alterations once critical loads are exceeded, establishing new rest configurations that defy prediction by elastic models alone. Third, damage evolution causes elastoplastic material parameters to drift over repeated loading cycles, demanding time‑dependent material descriptions. Conventional digital twin frameworks, which are typically designed for rigid bodies or linearly elastic materials, do not accommodate this triad of phenomena. Prior research has largely focused on either rigid dynamics\cite{anitescu1999time, lan2022affine, stramigioli2001modeling} or static elastic deformation\cite{modi2024simplicits, xie2025vid2sim, li2020incremental}, leaving the intertwined plasticity, nonlinear elasticity, and progressive damage of elastoplastic articulated objects (EAOs) unaddressed. As a result, robots frequently fail in tasks involving grasping, folding, or forecasting post‑interaction states, jeopardizing safety and reliability in both domestic and industrial contexts.

To overcome these limitations, we introduce BoxTwin, an interactive digital‑twin system specifically designed for EAOs. Instead of relying on hand‑crafted physics models with static parameters, BoxTwin incorporates a physics‑aware architecture that captures the full dynamics of such objects, including nonlinear elasticity, plastic yield thresholds, and damage progression, through calibrated constitutive models. The system continuously updates material properties to reflect time‑varying damage accumulation, enabling robust simulation and control of EAOs in real‑world robotic tasks. BoxTwin lays the groundwork for interactive digital twins that enable safe, adaptive robotic manipulation in real‑world environments.

\section{Method}
\subsection{Articulated Body}

An EAO is represented as a collection of rigid links that are connected by ideal hinge joints modeling the flexible creases of the structure. The topology is captured by a rooted kinematic tree
$$\mathcal{G}= (\mathcal{V},\mathcal{E});$$
a vertex $i\in\mathcal{V}$ denotes a link $\mathcal{L}_{i}$ equipped with a body‑fixed frame $\mathcal{F}_{i}$, while an edge $e=(p,i)\in\mathcal{E}$ corresponds to a hinge joint that allows a single relative rotation about a known axis $\mathbf{a}_{e}$. The joint angle $q_{e}\in\mathbb{R}$ is the only configuration variable associated with that edge.

The homogeneous transform that maps the world frame to link $i$ is obtained by composing the transforms along the unique path from the root to $i$:
$$\mathbf{T}_{0\to i}(\mathbf{q})=\mathbf{T}_{0\to p}\,\mathbf{T}_{p\to i}(q_{e}),$$
$$\mathbf{T}_{p\to i}(q_{e})=
\begin{bmatrix}
\mathbf R(\mathbf a_{e},q_{e}) &
\mathbf t_{p\!\to e}+ \mathbf R(\mathbf a_{e},q_{e})\,\mathbf t_{e\!\to i}\\[4pt]
\mathbf 0^{\!\top} & 1
\end{bmatrix},
$$
where $\mathbf{R}(\mathbf{a}_{e},q_{e})$ is the rotation produced by the hinge angle $q_{e}$ about axis $\mathbf{a}_{e}$, $\mathbf{t}_{p\!\to e}$ (expressed in $\mathcal{F}_{p}$) is the constant vector from the parent‑frame origin to the hinge point, and $\mathbf{t}_{e\!\to i}$ (expressed in the child frame after rotation) is the constant vector from the hinge point to the child‑frame origin.

Collecting all joint angles into the vector
$$\mathbf{q}= [q_{e_{1}},\dots ,q_{e_{|\mathcal{E}|}}]^{\!\top}$$
provides a compact description of the configuration.
Joint‑space dynamics follow the standard Lagrange formulation
$$\mathbf{M}(\mathbf{q})\ddot{\mathbf{q}}
+ \mathbf{C}(\mathbf{q},\dot{\mathbf{q}})\dot{\mathbf{q}}
+ \mathbf{g}(\mathbf{q})
= \boldsymbol{\tau},$$
where $\mathbf{M}(\mathbf{q})$ is the inertia matrix, $\mathbf{C}(\mathbf{q},\dot{\mathbf{q}})$ combines Coriolis and centrifugal effects, $\mathbf{g}(\mathbf{q})$ denotes the generalized gravity forces, and $\boldsymbol{\tau}$ collects the generalized forces generated by external contacts. 

\subsection{Elastic response of the creases}
The torque produced by a crease is highly nonlinear with respect to its bending deformation and cannot be faithfully captured by a simple linear spring.  Let $q_{e}$ be the instantaneous hinge angle and $\bar q_{e}$ the associated rest angle; the deformation that drives the elastic response is the scalar gap
$$\Delta_{e}=q_{e}-\bar q_{e}.$$
The elastic torque is introduced as a smooth, differentiable mapping
$$\tau^{\mathrm{el}}_{e}=f_{\theta_{e}}(\Delta_{e}),$$
where $f_{\theta_{e}}\colon\mathbb{R}\!\to\!\mathbb{R}$ is parameterised by a set of coefficients $\theta_{e}$ that are specific to edge $e$.  This formulation captures the strongly nonlinear hinge behaviour while remaining amenable to analytical differentiation, which is required for subsequent identification and optimization tasks.
\subsection{Plastic deformation}

Permanent set of a crease is modelled by allowing the rest angle $\bar q_{e}$ to evolve.  An internal scalar variable $\alpha_{e}$ records the accumulated plastic rotation, so that
$$\bar q_{e}=q^{0}_{e}+\alpha_{e},$$
with $q^{0}_{e}$ denoting the nominal angle. 
Plastic flow is activated once the elastic gap exceeds a material‑specific yield threshold $\Delta^{y}_{e}$.  The yield condition
$$\Phi_{e}(\Delta_{e})=|\Delta_{e}|-\Delta^{y}_{e}\le 0$$
defines an elastic regime ($\Phi_{e}\le 0$).  When $\Phi_{e}>0$ the internal variable evolves according to a smooth flow rule
$$\dot{\alpha}_{e}=g_{\psi_{e}}(\Delta_{e}),$$
where $g_{\psi_{e}}\colon\mathbb{R}\!\to\!\mathbb{R}$ is a function parameterised by coefficients $\psi_{e}$.

\subsection{Damage evolution}
Repeated folding induces micro‑fracture, fibre delamination and other degradation mechanisms, which manifest as a gradual loss of stiffness and a reduced capacity to accommodate permanent set\cite{linvill2017constitutive, coffin2017creasing}. To capture this effect a scalar damage variable $d_{e}\in[0,1]$ is introduced for each hinge; $d_{e}=0$ corresponds to an undamaged crease and $d_{e}=1$ to a mechanically inactive joint.  Damage attenuates both the elastic torque and the plastic flow:
$$\tau^{\mathrm{el}}_{e}=(1-d_{e})\,f_{\theta_{e}}\!\bigl(q_{e}-\bar q_{e}\bigr),\qquad
\dot{\alpha}_{e}=(1-d_{e})\,g_{\psi_{e}}\!\bigl(q_{e}-\bar q_{e}\bigr).$$
The multiplicative form preserves the sign and monotonicity of the underlying maps while ensuring a physically meaningful reduction of the joint’s load‑bearing capability.
The evolution of $d_{e}$ is driven by the cumulative amount of plastic slip that has occurred at the hinge.  Define the plastic‑slip accumulator
$$\mathcal{P}_{e}(t)=\int_{0}^{t}\! \bigl|\dot{\alpha}_{e}(\tau)\bigr|\,\mathrm{d}\tau,$$
which records the total absolute plastic rotation experienced up to time $t$.  The damage‑rate law is taken as a smooth, monotone function of this scalar measure,
$$\dot d_{e}=h_{\xi_{e}}\!\bigl(\mathcal{P}_{e}\bigr),$$
where $h_{\xi_{e}}:\mathbb{R}_{\ge0}\!\to\![0,1]$ is parameterised by a modest set of coefficients $\xi_{e}$.

\section{EVALUATION}

To assess the effectiveness of the proposed model, we carried out two experiments. In the first, we manually deformed an EAO composed of two panels and a single joint, keeping one panel anchored to the table. In the second experiment, we employed Aloha robotic arms to manipulate the EAOs and reproduced the entire sequence within a simulation environment. Both experiments were implemented using the MuJoCo simulator\cite{todorov2012mujoco}.

\subsection{Folding Test}

\begin{figure}[t]                       
    \centering
    \begin{subfigure}{\linewidth}
        \includegraphics[width=\linewidth]{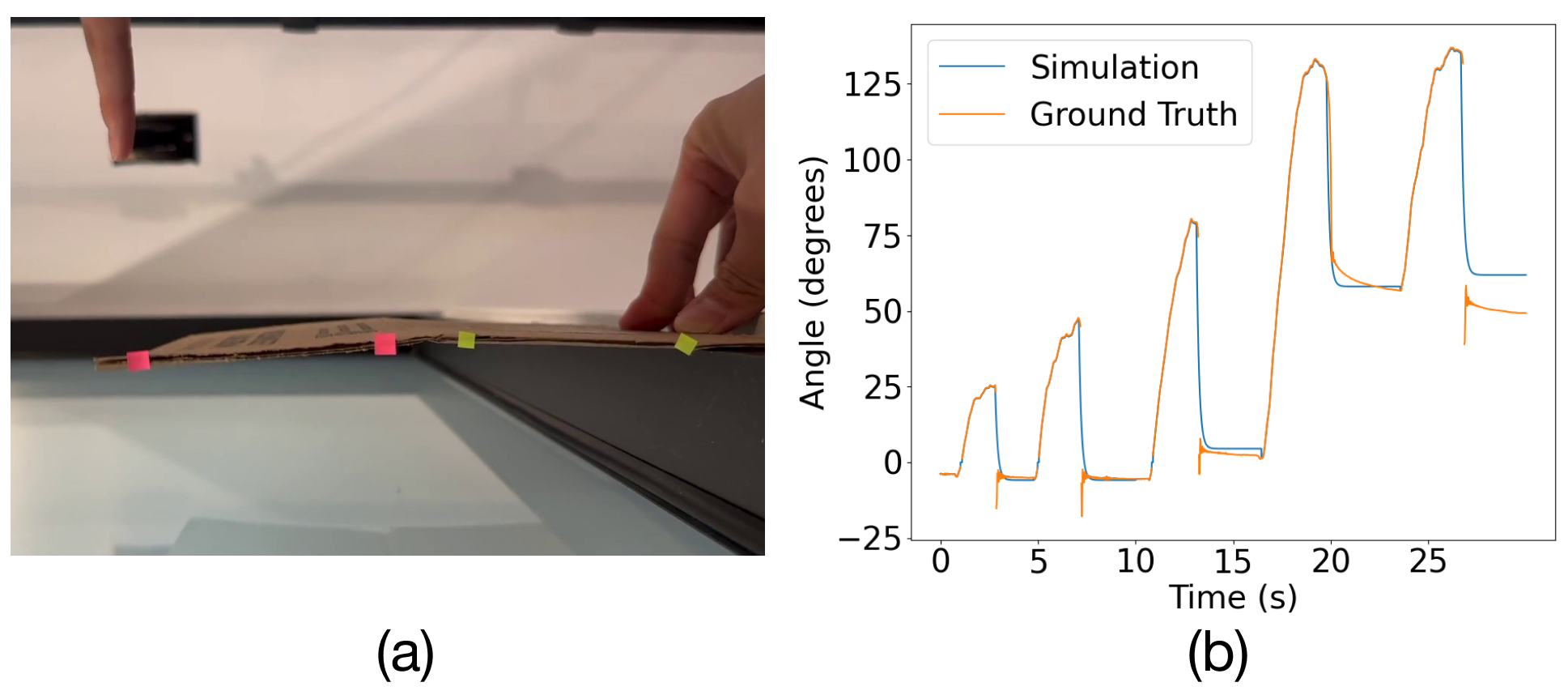}
        \label{fig:sub_a}
    \end{subfigure}
    \vspace{-6mm}
    \caption{Figure (a) depicts the experimental setup. Figure (b) showcases the expressive power of our proposed model in capturing the long‑horizon dynamics of the EAOs.}
    \label{fig:res1}
\end{figure}

To capture the joint angle of the EAO, color markers are affixed to the side of the panel; their positions are continuously monitored, allowing the joint angle to be derived as illustrated in Figure \ref{fig:res1}(a). Subsequently, the same actuation applied to the physical object is transferred in simulation, and the resulting joint angles are recorded throughout the motion. Figure \ref{fig:res1}(b) presents the simulated joint angle trajectory together with the corresponding measurements from the real‑world EAO.

Our proposed model captures the actual behavior of the object across the long sequence, faithfully reproducing both the instantaneous kinematic responses and the gradual evolution of material properties; it accurately tracks the joint‑angle trajectories despite the accumulation of hysteresis and wear, preserves the nuanced interplay between elastic, plastic, and damage mechanisms throughout repeated actuation cycles, and thereby provides a robust, high‑fidelity digital twin that remains reliable even as the system experiences prolonged exposure to complex loading patterns.

\subsection{Prediction Test}

\begin{figure}[t]                       
    \centering
    \begin{subfigure}{\linewidth}
        \includegraphics[width=\linewidth]{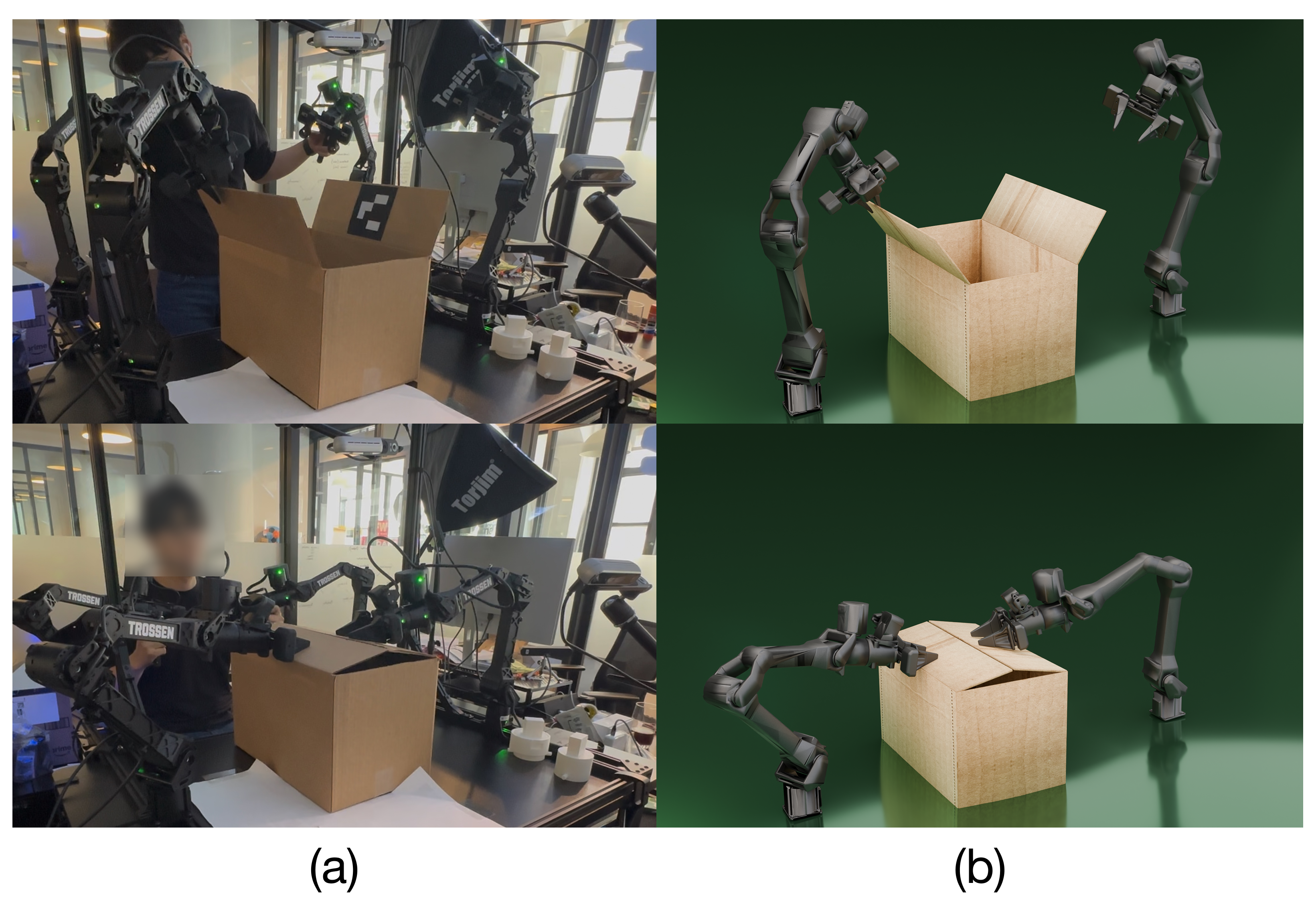}
    \end{subfigure}
    \vspace{-6mm}
    \caption{Figure (a) shows the frames captured from the real‑world video, and Figure (b) presents the corresponding simulated views.}
    \label{fig:res2}
\end{figure}

In this experiment we command the Trossen Aloha dual‑arm platform to interact with a variety of EAOs. The system operates in a leader–follower configuration: the leader arms receive the high‑level motion commands, while the follower arms execute the physical actions on the objects. Both simple EAOs that possess a single revolute joint and more complex assemblies that contain several coupled joints are handled by the same control framework. The follower arms are responsible for the complete manipulation sequence. They approach the object, apply the necessary forces to translate, rotate, or fold its individual links, and release it once the desired configuration is achieved.

After each trial we logged the full follower‑arm trajectory, including joint positions, velocities, torques and end‑effector pose at every step. Using these data we reconstructed an identical virtual scene in a physics‑based simulator: the Trossen Aloha dual‑arms were instantiated with the same kinematic and dynamic parameters (link lengths, masses, joint limits, friction), and the EAOs were modeled with the experimentally identified parameters. The recorded control signals were then replayed on the simulated follower arms, preserving the exact timing and magnitude of every command. By comparing the simulated joint angles and overall object configurations with the measurements obtained on the physical hardware, we observed a close correspondence across all test cases, from single‑joint modules to multi‑joint assemblies. Figure \ref{fig:res2} shows the replay results for a box with multiple joints. This validation demonstrates that the simulated environment can be reliably used for further algorithm development, predictive analysis, and transfer‑learning studies without the need for repetitive physical trials.

\section{CONCLUSIONS}

We introduce the first interactive digital‑twin framework that captures the full dynamics of elastoplastic articulated objects (EAOs) during real‑world robotic manipulation. By integrating a video‑driven reconstruction pipeline that simultaneously handles nonlinear elasticity, plastic deformation, and damage evolution, our physics‑aware twins predict post‑contact plastic behavior—something prior EAO models could not achieve. BoxTwin thus moves beyond static simulations, enabling adaptive, predictive control of deformable objects in unstructured environments where rigid‑body assumptions fail.

\addtolength{\textheight}{-12cm}   



\bibliographystyle{IEEEtran}
\bibliography{IEEEabrv, main.bib}

\begin{thebibliography}{10}
\providecommand{\url}[1]{#1}
\csname url@samestyle\endcsname
\providecommand{\newblock}{\relax}
\providecommand{\bibinfo}[2]{#2}
\providecommand{\BIBentrySTDinterwordspacing}{\spaceskip=0pt\relax}
\providecommand{\BIBentryALTinterwordstretchfactor}{4}
\providecommand{\BIBentryALTinterwordspacing}{\spaceskip=\fontdimen2\font plus
\BIBentryALTinterwordstretchfactor\fontdimen3\font minus \fontdimen4\font\relax}
\providecommand{\BIBforeignlanguage}[2]{{%
\expandafter\ifx\csname l@#1\endcsname\relax
\typeout{** WARNING: IEEEtran.bst: No hyphenation pattern has been}%
\typeout{** loaded for the language `#1'. Using the pattern for}%
\typeout{** the default language instead.}%
\else
\language=\csname l@#1\endcsname
\fi
#2}}
\providecommand{\BIBdecl}{\relax}
\BIBdecl

\bibitem{jiang2025phystwin}
H.~Jiang, H.-Y. Hsu, K.~Zhang, H.-N. Yu, S.~Wang, and Y.~Li, ``Phystwin: Physics-informed reconstruction and simulation of deformable objects from videos,'' \emph{arXiv preprint arXiv:2503.17973}, 2025.

\bibitem{malik2021digital}
A.~A. Malik and A.~Brem, ``Digital twins for collaborative robots: A case study in human-robot interaction,'' \emph{Robotics and Computer-Integrated Manufacturing}, vol.~68, p. 102092, 2021.

\bibitem{mu2025robotwin}
Y.~Mu, T.~Chen, Z.~Chen, S.~Peng, Z.~Lan, Z.~Gao, Z.~Liang, Q.~Yu, Y.~Zou, M.~Xu \emph{et~al.}, ``Robotwin: Dual-arm robot benchmark with generative digital twins,'' in \emph{Proceedings of the Computer Vision and Pattern Recognition Conference}, 2025, pp. 27\,649--27\,660.

\bibitem{feng2024pie}
Y.~Feng, Y.~Shang, X.~Li, T.~Shao, C.~Jiang, and Y.~Yang, ``Pie-nerf: Physics-based interactive elastodynamics with nerf,'' in \emph{Proceedings of the IEEE/CVF Conference on Computer Vision and Pattern Recognition}, 2024, pp. 4450--4461.

\bibitem{jiang2024vr}
Y.~Jiang, C.~Yu, T.~Xie, X.~Li, Y.~Feng, H.~Wang, M.~Li, H.~Lau, F.~Gao, Y.~Yang \emph{et~al.}, ``Vr-gs: A physical dynamics-aware interactive gaussian splatting system in virtual reality,'' in \emph{ACM SIGGRAPH 2024 Conference Papers}, 2024, pp. 1--1.

\bibitem{xie2024physgaussian}
T.~Xie, Z.~Zong, Y.~Qiu, X.~Li, Y.~Feng, Y.~Yang, and C.~Jiang, ``Physgaussian: Physics-integrated 3d gaussians for generative dynamics,'' in \emph{Proceedings of the IEEE/CVF Conference on Computer Vision and Pattern Recognition}, 2024, pp. 4389--4398.

\bibitem{chen2022virtual}
H.-y. Chen, E.~Tretschk, T.~Stuyck, P.~Kadlecek, L.~Kavan, E.~Vouga, and C.~Lassner, ``Virtual elastic objects,'' in \emph{Proceedings of the IEEE/CVF Conference on Computer Vision and Pattern Recognition}, 2022, pp. 15\,827--15\,837.

\bibitem{li2023pac}
X.~Li, Y.-L. Qiao, P.~Y. Chen, K.~M. Jatavallabhula, M.~Lin, C.~Jiang, and C.~Gan, ``Pac-nerf: Physics augmented continuum neural radiance fields for geometry-agnostic system identification,'' \emph{arXiv preprint arXiv:2303.05512}, 2023.

\bibitem{qiao2022neuphysics}
Y.-L. Qiao, A.~Gao, and M.~Lin, ``Neuphysics: Editable neural geometry and physics from monocular videos,'' \emph{Advances in Neural Information Processing Systems}, vol.~35, pp. 12\,841--12\,854, 2022.

\bibitem{zhang2024physdreamer}
T.~Zhang, H.-X. Yu, R.~Wu, B.~Y. Feng, C.~Zheng, N.~Snavely, J.~Wu, and W.~T. Freeman, ``Physdreamer: Physics-based interaction with 3d objects via video generation,'' in \emph{European Conference on Computer Vision}.\hskip 1em plus 0.5em minus 0.4em\relax Springer, 2024, pp. 388--406.

\bibitem{zhong2024reconstruction}
L.~Zhong, H.-X. Yu, J.~Wu, and Y.~Li, ``Reconstruction and simulation of elastic objects with spring-mass 3d gaussians,'' in \emph{European Conference on Computer Vision}.\hskip 1em plus 0.5em minus 0.4em\relax Springer, 2024, pp. 407--423.

\bibitem{anitescu1999time}
M.~Anitescu, F.~A. Potra, and D.~E. Stewart, ``Time-stepping for three-dimensional rigid body dynamics,'' \emph{Computer methods in applied mechanics and engineering}, vol. 177, no. 3-4, pp. 183--197, 1999.

\bibitem{lan2022affine}
L.~Lan, D.~M. Kaufman, M.~Li, C.~Jiang, and Y.~Yang, ``Affine body dynamics: Fast, stable \& intersection-free simulation of stiff materials,'' \emph{arXiv preprint arXiv:2201.10022}, 2022.

\bibitem{stramigioli2001modeling}
S.~Stramigioli, \emph{Modeling and IPC control of interactive mechanical systems—A coordinate-free approach}.\hskip 1em plus 0.5em minus 0.4em\relax Springer, 2001.

\bibitem{modi2024simplicits}
V.~Modi, N.~Sharp, O.~Perel, S.~Sueda, and D.~I. Levin, ``Simplicits: Mesh-free, geometry-agnostic elastic simulation,'' \emph{ACM Transactions on Graphics (TOG)}, vol.~43, no.~4, pp. 1--11, 2024.

\bibitem{xie2025vid2sim}
Z.~Xie, Z.~Liu, Z.~Peng, W.~Wu, and B.~Zhou, ``Vid2sim: Realistic and interactive simulation from video for urban navigation,'' in \emph{Proceedings of the Computer Vision and Pattern Recognition Conference}, 2025, pp. 1581--1591.

\bibitem{li2020incremental}
M.~Li, Z.~Ferguson, T.~Schneider, T.~R. Langlois, D.~Zorin, D.~Panozzo, C.~Jiang, and D.~M. Kaufman, ``Incremental potential contact: intersection-and inversion-free, large-deformation dynamics.'' \emph{ACM Trans. Graph.}, vol.~39, no.~4, p.~49, 2020.

\bibitem{linvill2017constitutive}
E.~Linvill, M.~Wallmeier, and S.~{\"O}stlund, ``A constitutive model for paperboard including wrinkle prediction and post-wrinkle behavior applied to deep drawing,'' \emph{International Journal of Solids and Structures}, vol. 117, pp. 143--158, 2017.

\bibitem{coffin2017creasing}
D.~W. Coffin and M.~Nyg{\aa}rds, ``Creasing and folding,'' in \emph{Transactions of the 16th fundamental research symposium}, 2017, pp. 69--136.

\bibitem{todorov2012mujoco}
E.~Todorov, T.~Erez, and Y.~Tassa, ``Mujoco: A physics engine for model-based control,'' in \emph{2012 IEEE/RSJ International Conference on Intelligent Robots and Systems}.\hskip 1em plus 0.5em minus 0.4em\relax IEEE, 2012, pp. 5026--5033.

\end{thebibliography}

\end{document}